\documentclass{article}

\usepackage[final]{neurips_2025}

\usepackage[utf8]{inputenc} % allow utf-8 input
\usepackage[T1]{fontenc}    % use 8-bit T1 fonts
\usepackage{hyperref}       % hyperlinks
\usepackage{url}            % simple URL typesetting
\usepackage{booktabs}       % professional-quality tables
\usepackage{amsfonts}       % blackboard math symbols
\usepackage{nicefrac}       % compact symbols for 1/2, etc.
\usepackage{microtype}      % microtypography
\usepackage{xcolor}         % colors
\usepackage{amsmath}
\usepackage{graphicx}
\usepackage{enumitem}
\usepackage{wrapfig}
\usepackage{multirow}
\usepackage{marvosym}
\usepackage{amsmath}
\usepackage{graphicx}
\usepackage{enumitem}
\usepackage{wrapfig}
\usepackage{multirow}
\usepackage{booktabs}
\usepackage{pifont}
\usepackage{chngcntr}
\usepackage{marvosym}
\usepackage{algorithm}
\usepackage{algpseudocode}
\usepackage{amsmath}
\title{AlignedGen: Aligning Style Across Generated Images}

\author{\hspace{-22.5pt}
Jiexuan Zhang\textsuperscript{*},\quad
Yiheng Du\textsuperscript{*},\quad Qian Wang,\quad
Weiqi Li,\quad Yu Gu,\quad Jian Zhang\textsuperscript{\Letter} \\[5pt]
School of Electronic and Computer Engineering, Peking University \\[5pt]
\url{https://github.com/Jiexuanz/AlignedGen}
}

\begin{document}

\renewcommand*{\thefootnote}{*}
\footnotetext[1]{Equal Contribution. \Letter: Corresponding author, zhangjian.sz@pku.edu.cn.}
% \renewcommand*{\thefootnote}{\Letter}
% \footnotetext[1]{Corresponding author, zhangjian.sz@pku.edu.cn.}

\maketitle

\begin{figure}[h]
\vspace{-0pt}
\begin{center}
\includegraphics[width=1.00\linewidth]{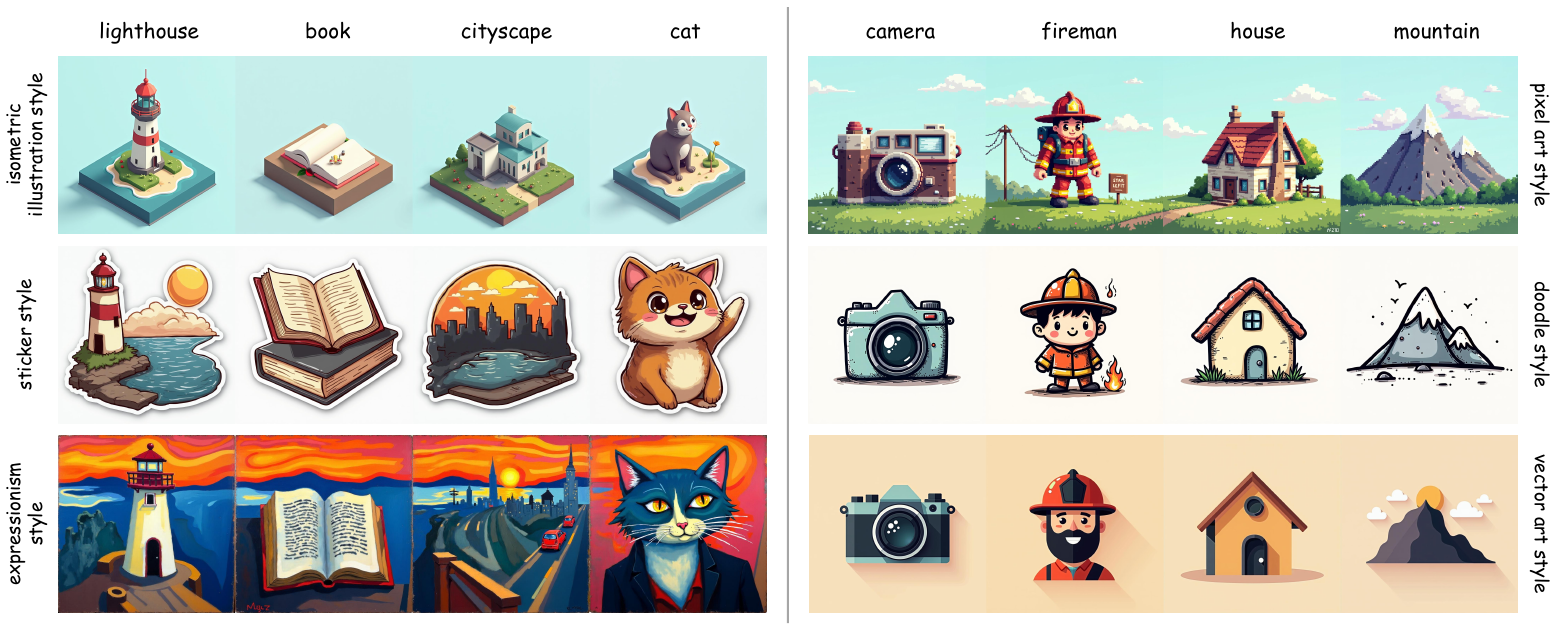}
\end{center}
\vspace{-5pt}
\caption{\textbf{Generation results of AlignedGen.} As a training-free method, AlignedGen generates images with varying content while ensuring high style consistency.}
\vspace{0pt}
\label{fig:abs}
\end{figure}

\begin{abstract}
Despite their generative power, diffusion models struggle to maintain style consistency across images conditioned on the same style prompt, hindering their practical deployment in creative workflows.
While several training-free methods attempt to solve this, they are constrained to the U-Net architecture, which not only leads to low-quality results and artifacts like object repetition but also renders them incompatible with superior Diffusion Transformer (DiT).
To address these issues, we introduce \textit{\textbf{AlignedGen}}, a novel training-free framework that enhances style consistency across images generated by DiT models.
Our work first reveals a critical insight: naive attention sharing fails in DiT due to conflicting positional signals from improper position embeddings.
We introduce Shifted Position Embedding (ShiftPE), an effective solution that resolves this conflict by allocating a non-overlapping set of positional indices to each image.
Building on this foundation, we develop Advanced Attention Sharing (AAS), a suite of three techniques meticulously designed to fully unleash the potential of attention sharing within the DiT.
Furthermore, to broaden the applicability of our method, we present an efficient query, key, and value feature extraction algorithm, enabling our method to seamlessly incorporate external images as style references. 
Extensive experimental results validate that our method effectively enhances style consistency across generated images while maintaining precise text-to-image alignment.
\end{abstract}

\section{Introduction}
\label{sec:intro}

\begin{figure*}[!t]
\vspace{0pt}
\begin{center}
\includegraphics[width=1.00\linewidth]{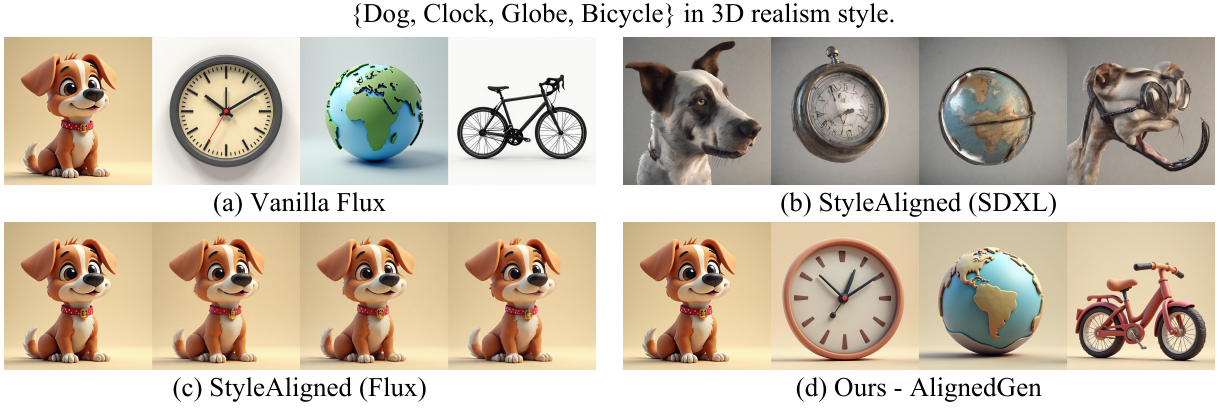}
\end{center}
\vspace{-10pt}
\caption{\textbf{Qualitative comparison with Vanilla Flux and StyleAligned}.
(a) Vanilla Flux produces stylistically incoherent results.
(b) StyleAligned (SDXL) suffers from poor image quality and exhibits severe artifacts.
(c) StyleAligned (Flux) suffers from complete mode collapse, yielding identical outputs that disregard the textual prompt.
(d) In contrast, our method maintains style consistency across images while faithfully adhering to all four concepts in the prompt.}
\vspace{-10pt}
\label{fig:intro}
\end{figure*}

The advent of large-scale text-to-image diffusion models~\citep{ho2020denoising, rombach2022high, podell2023sdxl, esser2024scaling, flux} marks a paradigm shift in generative AI, offering unprecedented capabilities for synthesizing visual content from textual descriptions. 
These models have evolved from U-Net~\citep{ronneberger2015u} to Diffusion Transformer (DiT) — the backbone of state-of-the-art diffusion models such as Flux and Qwen-Image. 
Despite these rapid advancements, all diffusion models struggle to maintain high style consistency across a series of generated images solely via style text description, as shown in Fig.~\ref{fig:intro} (a). % 可以不要也可以要
This inconsistency severely hinders their practical deployment in content creation, such as illustrating books and graphic novels, designing cohesive sets of virtual assets, or producing datasets for downstream tasks.

% An intuitive approach to this challenge would be to first generate a single style reference image and then apply existing style image generation techniques.
One might consider a seemingly straightforward approach to address this challenge: generate a single style reference image and then leverage existing style image generation techniques.
However, existing style image generation techniques all have shortcomings and defects, which limits the performance of this approach.
On one hand, tuning-based style image generation methods~\citep{hu2021lora, kumari2023multi, ruiz2023dreambooth, gal2022image, frenkel2024implicit, shah2024ziplora} like LoRA offer exceptional style fidelity but require per-style optimization, rendering them impractical for real-time use. 
On the other hand, dataset-driven style image generation methods~\citep{qi2024deadiff, gao2024styleshot, xing2024csgo, liu2023stylecrafter, huang2025artcrafter} promise zero-shot generalization but frequently fail to capture the novel, unseen styles. 
Consequently, users are trapped in a dilemma: either endure the slow, costly process of optimization for quality or settle for the limited and often unreliable generalization of data-driven methods.

StyleAligned~\citep{hertz2024style}, a training-free approach for style-aligned image generation, has emerged to address these limitations.
While promising, its reliance on the aging SDXL architecture inherently limits its output quality, often producing artifacts (Fig.~\ref{fig:intro} (b)).
Most critically, its core strategy is architecturally incompatible with DiT – the superior generative architecture for diffusion models. 
Designed explicitly for U-Net’s self-attention, it fails when applied to DiT, resulting in loss of text control and severe content collapse, as shown in Fig.~\ref{fig:intro} (c). 
% This reveals a crucial, unaddressed gap: how to achieve robust, training-free style-aligned image generation for the superior DiT architecture.
This reveals a crucial, unaddressed gap – one that our work addresses: how to achieve robust, training-free style-aligned image generation for DiT-based diffusion models.

To address the limitations of existing methods, we propose \textit{\textbf{AlignedGen}}, a novel framework to enhance style consistency across images generated by DiT-based diffsuion model like Flux.
First, we conduct a critical investigation into a fundamental problem: the failure of attention sharing mechanisms in the DiT. 
We identify the root cause as an improper setup of Rotary Position Embedding (RoPE), which creates conflicting positional signals.
To resolve this, we propose \textbf{Shift}ed \textbf{P}osition \textbf{E}mbedding (\textbf{ShiftPE}), a novel and elegant solution that assigns non-overlapping positional spaces to each image. 
This not only fixes generation collapse and restores text controllability but also offers a crucial insight for all future attention sharing works on DiT.
Building upon this breakthrough, we further develop \textbf{A}dvanced \textbf{A}ttention \textbf{S}haring (\textbf{AAS}), a suite of three specialized techniques - Selective Attention Sharing, Controllable Style Consistency Via Key Scaling, and Layer-Selective Application.
These techniques are specifically designed for the DiT architecture, further unleashing its generative potential and leading to superior output quality.
Finally, to extend our method's utility, we introduce an approximate query, key, value extraction algorithm, allowing it to condition on external images as style references and expanding its application scope.
% Crucially, AlignedGen is designed to be plug-and-play and training-free, facilitating immediate integration with established tools like ControlNet~\cite{zhang2023adding} and DreamBooth. 
Extensive experiments demonstrate that our method significantly outperforming prior work in both style consistency and faithfulness to text prompts.
In summary, our contributions are as follows:

\noindent \ding{113}~(1) We present AlignedGen, the pioneering training-free style-aligned image generation framework designed for DiT. AlignedGen operates without any fine-tuning or extra modules, enabling the efficient utilization of large DiT models with billions of parameters.

\noindent \ding{113}~(2) We discover that conflicting position embeddings are the root cause of attention sharing failure in DiT. We introduce Shifted Position Embedding (ShiftPE), a novel and effective solution that enables viable attention sharing in DiT architecture for the first time.

\noindent \ding{113}~(3) We propose Advanced Attention Sharing (AAS), a set of three specialized techniques engineered to unleash the full potential of DiT for style-aligned generation, markedly improving style fidelity and image coherence.

\noindent \ding{113}~(4) We devise an algorithm to extract approximate query, key, and value features from any external image. This allows our framework to support user-provided images as style references without retraining, significantly broadening its applicability.

% \noindent \ding{113}~(5) Extensive experiments demonstrate that our method generates high-quality images with superior style consistency and accurate alignment with text prompts.

\section{Related Work}
\label{sec:related}
\subsection{Attention Control in Diffusion Models}
Controlling the generative process of diffusion models~\citep{ho2020denoising, nichol2021glide, saharia2022photorealistic, rombach2022high, podell2023sdxl, esser2024scaling, flux,wang2024360dvd} via attention manipulation~\citep{cao2023masactrl, mo2024freecontrol, lin2024ctrl,li2024omnidrag,wang2025mind,mou2024t2i} has become a prominent research direction. 
Pioneering works like Prompt-to-Prompt~\citep{hertz2022prompt} and Plug-and-Play~\citep{tumanyan2023plug} demonstrated that by editing cross-attention or injecting self-attention features, one could precisely control image content and layout. 
Crucially, these methods are designed for earlier architectures that explicitly separated self-attention from cross-attention.
The advent of state-of-the-art, Transformer-based models like DiT~\citep{peebles2023scalable} and Flux has challenged this paradigm by unifying both functions into a single, multi-modal attention mechanism. 
Consequently, a gap has emerged for attention-based control techniques that are compatible with these modern, unified architectures. 
Our work aims to bridge this gap.

\subsection{Style Image Generation}
Classic neural style transfer methods~\citep{zhang2013style, gatys2016image, an2021artflow, deng2022stytr2, huang2017arbitrary, li2017universal, liu2021adaattn, zhang2022domain} excelled at recomposing a single content image in the style of another. 
Style image generation has moved beyond the one-to-one paradigm of traditional style transfer, now focusing on the scalable generation of images from a single style reference across diverse prompts. 
Current diffusion-based methods for this task largely fall into two categories.
Tuning-based methods~\citep{hu2021lora, kumari2023multi, ruiz2023dreambooth, gal2022image, frenkel2024implicit, shah2024ziplora} like LoRA and DreamBooth achieve high fidelity to a target style but require per-style optimization. 
Conversely, dataset-driven models~\citep{qi2024deadiff, gao2024styleshot, xing2024csgo, liu2023stylecrafter, huang2025artcrafter} are trained on large style collections but struggle with the cost of dataset curation and may not generalize to unseen styles.
A distinct, yet related, challenge is ensuring style consistency across generated images from the same style prompt. 
StyleAligned~\citep{hertz2024style} addresses this by sharing information between self-attention layers in a tuning-free manner. 
However, StyleAligned's efficacy is confined to U-Net, and it fails to generalize to DiT. 
Our work addresses this critical architectural gap with an advanced attention sharing module designed for DiT.

\section{Method}
\label{sec:method}

\begin{figure*}[!t]
\vspace{0pt}
\centering
\includegraphics[width=1.00\textwidth]{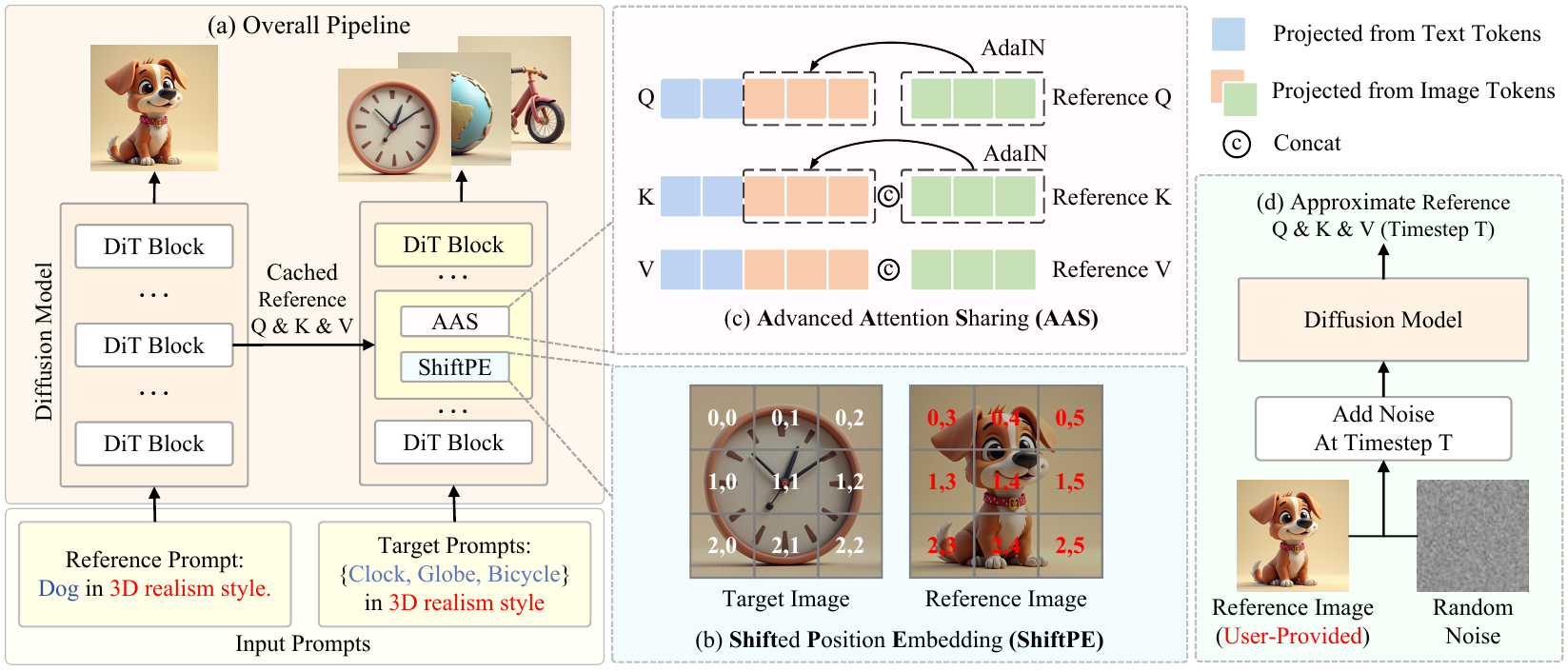}
\vspace{-17.5pt}
\caption{
\textbf{Our proposed framework}. 
(a) Overall pipeline, where our ShiftPE and AAS module are integrated into specific layers of DiT, replacing the MM-Attention. 
(b)(c) Detailed illustrations of the ShiftPE and AAS, respectively. 
(d) Procedure for extracting features from a user-provided style reference image, which serve as input to the AAS module.}
\label{fig:pipe}
\vspace{0pt}
\end{figure*}

Our goal is to generate an image set \( \mathcal{I}\) that is stylistically consistent and faithful to the corresponding set of prompts \( \mathcal{T}\).
To achieve this, we employ a reference-target strategy, illustrated in Fig.~\ref{fig:pipe} (a), we designate one image as the style reference \( I^{ref} \), and constrain the remaining target images \( I^{tar} \) to match its style. 
This distinction is maintained at the feature level, where we differentiate between reference (\(\mathbf{Q}^{ref}\), \(\mathbf{K}^{ref}\), \(\mathbf{V}^{ref}\)) and targets (\(\mathbf{Q}^{tar}\), \(\mathbf{K}^{tar}\), \(\mathbf{V}^{tar}\)) representations.

\subsection{Preliminary}
\paragraph{Diffusion Transformer (DiT).} 
The Diffusion Transformer~\citep{peebles2023scalable} adapts the Transformer architecture for generative modeling, operating on sequences of latent image tokens $\textbf{X}\in \mathbb{R}^{N\times d}$ and text condition tokens $\textbf{C} \in \mathbb{R}^{M\times d}$.
These tokens are typically augmented with positional information via Rotary Position Embeddings (RoPE)~\citep{su2024roformer}.
Instead of traditional self-attention and cross-attention, modern DiT models such as Flux~\citep{flux} employ a unified Multi-Modal Attention (MM-Attention) mechanism. 
In this paradigm, both image and text tokens are projected through a set of projection matrices to obtain the corresponding representations of $\mathbf{Q}_{\{img,txt\}}$, $\mathbf{K}_{{\{img,txt\}}}$ and $\mathbf{V}_{{\{img,txt\}}}$. This process is formulated as follows:
\begin{equation}
\begin{aligned}
\label{eq:proj}
& \mathbf{\{Q,K,V\}}_{img} = \mathbf{W}_{img}^{\{Q,K,V\}} \mathbf{X}_{}, \ \mathbf{\{Q,K,V\}}_{img}\in\mathbb{R}^{N\times d_k}& \\
& \mathbf{\{Q,K,V\}}_{txt} = \mathbf{W}_{txt}^{\{Q,K,V\}} \mathbf{C}_{}, \ \mathbf{\{Q,K,V\}}_{txt}\in\mathbb{R}^{M\times d_k}& 
\end{aligned}
\end{equation}
where $\mathbf{W}_{\{img,\ txt\}}^{\{Q,K,V\}}$ denote the projection matrices, $d_k$ is the embedding dimension in attention.
The $\mathbf{Q}_{\{img,txt\}}$, $\mathbf{K}_{{\{img,txt\}}}$ and $\mathbf{V}_{{\{img,txt\}}}$ are concatenated along the sequence length dimension to obtain the complete $\mathbf{Q}\in\mathbb{R}^{(M+N)\times d_k}$, $\mathbf{K}\in\mathbb{R}^{(M+N)\times d_k}$, and $\mathbf{V}\in\mathbb{R}^{(M+N)\times d_k}$.
After that, the final output is computed using the following formula:
\begin{equation}
\begin{aligned}
\label{eq:attn}
& \text{Attention}(\mathbf{Q}, \mathbf{K}, \mathbf{V}) = \text{Softmax}\left(\frac{\mathbf{Q}\mathbf{K}^{\mathrm{T}}}{\sqrt{d_k}}\right)\mathbf{V}.
\end{aligned}
\end{equation}

% \begin{equation}
% \begin{aligned}
% \label{eq:concat}
% \mathbf{Q} = \text{Concat}(\mathbf{Q}_{txt}, \mathbf{Q}_{img}),\ \mathbf{K} = \text{Concat}(\mathbf{K}_{txt}, \mathbf{K}_{img}),\ \mathbf{V} = \text{Concat}(\mathbf{V}_{txt}, \mathbf{V}_{img}).
% \end{aligned}
% \end{equation}

\paragraph{Naive Attention Sharing.}
Attention sharing, the core technique of the style-aligned image generation method~\citep{hertz2024style} in U-Net architecture~\citep{ronneberger2015u}, can be straightforwardly adapted to MM-Attention within DiT models.
We refer to this straightforward adaptation as Naive Attention Sharing. 
Specifically, it aligns $\mathbf{Q}^{tar}$ with $\mathbf{Q}^{ref}$ through Adaptive Instance Normalization (AdaIN)~\citep{huang2017arbitrary} to form the final query $\mathbf{Q}^{F}$. 
Similarly, it uses AdaIn to align the target key $\mathbf{K}^{tar}$ with $\mathbf{K}^{ref}$ to form $\mathbf{\hat{K}}^{tar}$.
Then, the final key $\mathbf{K}^{F}$ and value $\mathbf{V}^{F}$ are formed by concatenating the target features $\mathbf{\hat{K}}^{tar}$, $\mathbf{V}^{tar}$ with the reference features $\mathbf{K}^{ref}$, $\mathbf{V}^{ref}$ along the sequence dimension. 
This process is formulated as follows:
\begin{equation}
\begin{aligned}
\label{eq:naive}
& \text{AdaIN}(x, y) = \sigma(y) \left( \dfrac{x - \mu(x)}{\sigma(x)} \right) + \mu(y), \\
& \mathbf{Q}^{F} = \text{AdaIN}(\mathbf{Q}^{tar}, \mathbf{Q}^{ref})\in\mathbb{R}^{(M+N)\times d_k}, \\
& \mathbf{K}^{F} = \text{Concat}(\mathbf{\hat{K}}^{tar}, \mathbf{K}^{ref})\in\mathbb{R}^{2(M+N)\times d_k}, \ \text{where} \ \mathbf{\hat{K}}^{tar}=\text{AdaIN}(\mathbf{K}^{tar}, \mathbf{K}^{ref}), \\
& \mathbf{V}^{F} = \text{Concat}(\mathbf{V}^{tar}, \mathbf{V}^{ref})\in\mathbb{R}^{2(M+N)\times d_k}, \\
\end{aligned}
\end{equation}
with $\mu(\cdot)$ and $\sigma(\cdot)$ representing the mean and standard deviation.
In the target images $I^{tar}$ generation process, the query, key, and value in MM-Attention are replaced by $\mathbf{Q}^{F}$, $\mathbf{K}^{F}$, and $\mathbf{V}^{F}$, respectively.
% with the results computed according to Equation \ref{eq:attn}.

% We visualize the attention map originating from a query point (red box) in the target image. 
% Default RoPE exhibits a strong spatial bias, rigidly constraining attention to the same coordinates in the reference image, leading to undesirable content leakage.
% In contrast, ShiftPE breaks this rigid spatial dependency by introducing shifted positional indices, allowing the model to attend to broader, semantically relevant regions (e.g., the periphery of the snowy landscape).
% This is confirmed quantitatively, where the right plot shows ShiftPE's attention distributed by L1 distance, unlike RoPE's highly localized peak.
\begin{figure*}[!t]
\vspace{0pt}
\centering
\includegraphics[width=1.00\textwidth]{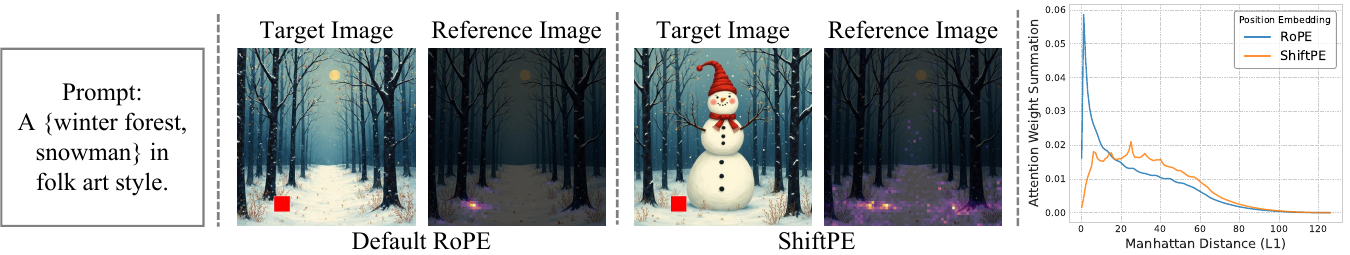}
\vspace{-15pt}
\caption{
\textbf{ShiftPE decouples attention from spatial location to enable semantic correspondence}. 
We visualize attention map originating from a query point (red box). 
(a) RoPE exhibits a strong spatial bias, rigidly constraining attention to the same coordinates in the reference image and causing content leakage. 
(b) ShiftPE breaks this spatial dependency, allowing attention to focus on broader, semantically relevant regions (e.g., the surrounding snowy landscape). 
(c) The plot of attention weight summed by L1 distance quantitatively confirms this: ShiftPE distributes attention across a wider area, whereas RoPE's attention is sharply localized at the query's exact coordinates.
}
\label{fig:shiftpe}
\vspace{-15pt}
\end{figure*}

\subsection{The limitation of Naive Attention Sharing in DiT}

Applying Naive Attention Sharing to DiT, while seemingly a straightforward strategy, results in a notable collapse in performance. 
As illustrated in Fig.~\ref{fig:intro} (c), this direct approach severely degrades text controllability and causes prominent content leakage from the reference image, leading to a catastrophic failure mode where all generated images become mere copies of the reference.
This raises a critical question: \textbf{Why does Naive Attention Sharing fail so markedly within the DiT}?

To answer this, we conduct a systematic investigation into the interplay between attention sharing mechanism and the DiT. 
Our investigation reveals that the root cause is not a flaw in the attention sharing mechanism per se, but a subtle yet critical issue we term \textbf{Positional Collision}.
While RoPE provides unique spatial coordinates $(i,j)$ for each token within a single image's $h\times w$ latent grid, the naive sharing scheme inadvertently assigns identical positional embeddings to tokens at the same coordinates $(i,j)$ in both the reference and target images. 
This collision induces a erroneous bias, forcing tokens to disproportionately attend to their spatial counterparts across images, irrespective of content or textual guidance, as illustrated in Fig.~\ref{fig:shiftpe} (RoPE).
To resolve this, we introduce a simple yet effective solution derived from this insight: \textbf{Shifted} \textbf{P}osition \textbf{E}mbedding (\textbf{ShiftPE}). 
The core idea is to ensure the reference and target images occupy distinct, non-overlapping coordinate spaces. We achieve this by virtually placing the reference image ``next to" the target images, remapping its positional indices from $(i,j)$ to $(i,j+w)$, a visual depiction is provided in Fig.~\ref{fig:pipe} (b).
As shown in Fig.~\ref{fig:shiftpe} (ShiftPE), ShiftPE resolves the positional collision, decoupling the spatial representations and thereby eliminating the content leakage.
To quantitatively validate this, we plot the sum of attention weights as a function of L1 distance from query point in Fig.~\ref{fig:shiftpe}.
The plot confirms that while RoPE's attention is sharply localized, ShiftPE enables the model to distribute attention over a much wider, semantically relevant area.
This frees the attention mechanism to focus on regions of high semantic and stylistic relevance, rather than being constrained by rigid spatial correspondence.

\subsection{Advanced Attention Sharing}
While ShiftPE effectively resolves the problem of positional collision in Naive Attention Sharing, unlocking state-of-the-art performance demands a solution that works well with the specifics of the DiT architecture. 
The DiT components are not inherently designed for the attention sharing.
Therefore, we introduce a series of advanced modifications that enable smooth, effective cooperation with DiT’s structure, significantly enhancing style consistency and generation quality.

\paragraph{Selective Attention Sharing (SAS).}
Directly applying full attention sharing of key and value in MM-Attention has a critical drawback.
In MM-Attention, keys and values are derived from both image and text tokens. 
Consequently, full attention sharing leads to prompt contamination: textual information from the reference prompt leaks into the generation process of the target image.
This contamination can disrupt the conditioning provided by the target prompt, leading to degraded alignment between text and image.
To address this, we introduce Selective Attention Sharing, which shares only key and value originating from image tokens, shown in Fig.~\ref{fig:pipe} (c).
This strategy not only effectively enhances the alignment between images and their corresponding prompts but also reduces computational overhead.
Finally, the $\mathbf{Q}^{F}$, $\mathbf{K}^{F}$ and $\mathbf{V}^{F}$ can be expressed as follow:
\begin{equation}
\begin{aligned}
\label{eq:advanced}
& \mathbf{Q}^{F} = \text{Concat}(\mathbf{Q}^{tar}_{txt},\ \mathbf{\hat{Q}}^{tar}_{img} )\in\mathbb{R}^{(M+N)\times d_k},\ \text{where} \ \mathbf{\hat{Q}}^{tar}_{img} = \text{AdaIN}(\mathbf{Q}^{tar}_{img}, \mathbf{Q}^{ref}_{img}), \\
& \mathbf{K}^{F} = \text{Concat}(\mathbf{K}^{tar}_{txt},\ \mathbf{\hat{K}}^{tar}_{img}, \mathbf{K}^{ref}_{img})\in\mathbb{R}^{(M+2N)\times d_k},\text{where} \ \mathbf{\hat{K}}^{tar}_{img}=\text{AdaIN}(\mathbf{K}^{tar}_{img}, \mathbf{K}^{ref}_{img}), \\
& \mathbf{V}^{F} = \text{Concat}(\mathbf{V}^{tar}_{txt},\  \mathbf{V}^{tar}_{img},\ \mathbf{V}^{ref}_{img})\in\mathbb{R}^{(M+2N)\times d_k}. \\
\end{aligned}
\end{equation}

\begin{figure*}[!t]
\vspace{0pt}
\centering
\includegraphics[width=0.90\textwidth]{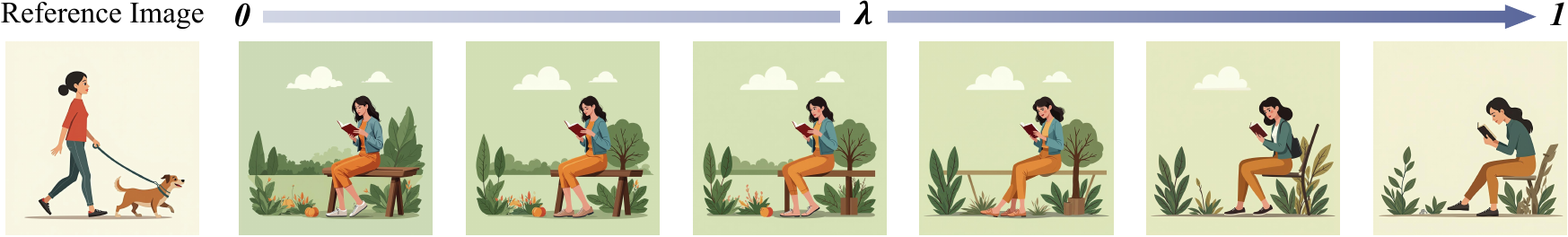}
\vspace{0pt}
\caption{The impact of scaling factor $\lambda$. As it increases, the target image progressively aligns with the style of
the reference image.}
\label{fig:lambda}
\vspace{0pt}
\end{figure*}

\begin{figure*}[!t]
\vspace{0pt}
\centering
\includegraphics[width=0.90\textwidth]{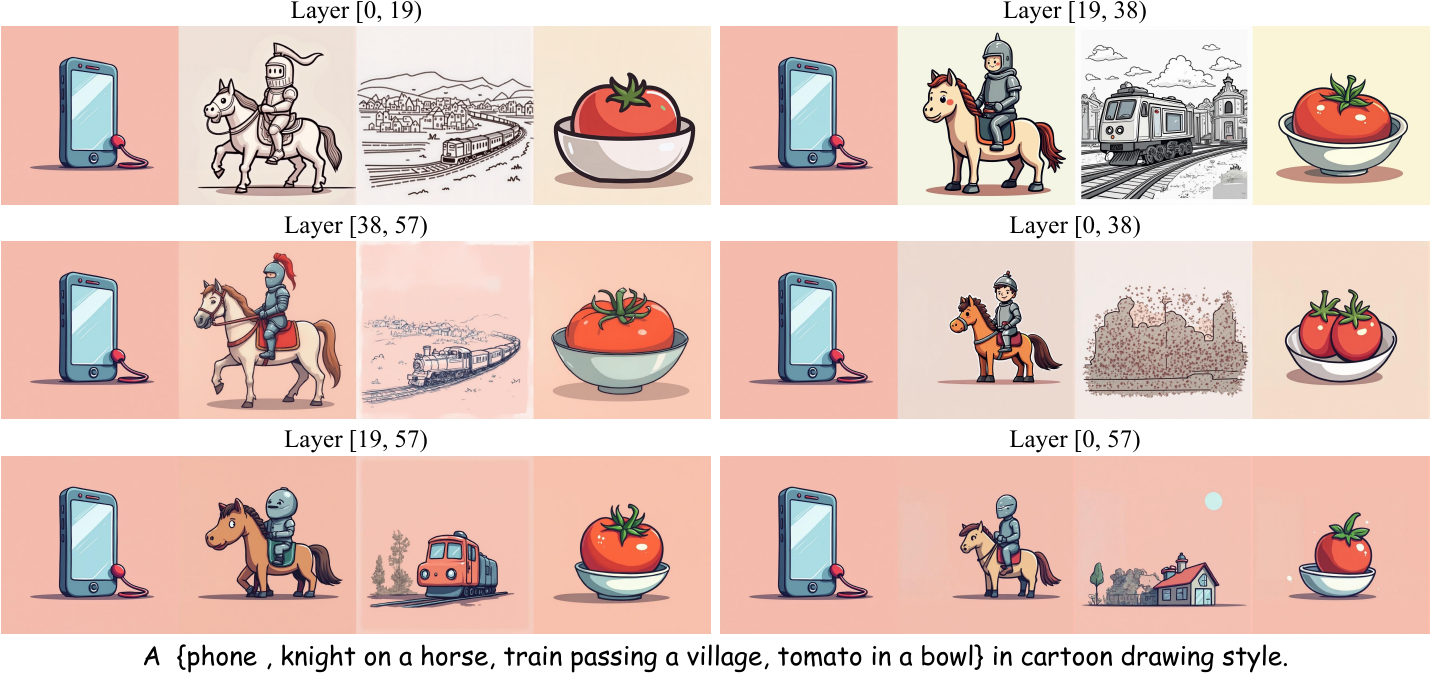}
\vspace{0pt}
\caption{Visualization of results generated by applying our method in different layers of Flux.}
\label{fig:block}
\vspace{-5pt}
\end{figure*}

\paragraph{Controllable Style Consistency Via Key Scaling.}
To provide explicit control over the strength of style consistency, we introduce a simple yet effective modulation parameter, $\lambda$.
Our approach scales the $\mathbf{K}_{img}^{ref}$ derived from the reference image tokens, before it is used in the attention sharing mechanism.
Intuitively, this scaling directly influences the attention scores. 
A larger $\lambda$ increases the magnitude of the dot-product similarities between the target query and the reference key. This, in turn, assigns higher attention weights to the reference style during the softmax operation, resulting in stronger style consistency, as shown in Fig.~\ref{fig:lambda}.
The final $\mathbf{K}^{F}$ is defined as:
\begin{equation}
\begin{aligned}
\label{eq:strength}
\mathbf{K}^{F} = \text{Concat}(\mathbf{K}_{txt}^{tar}, \hat{\mathbf{K}}_{img}^{tar}, \lambda \cdot \mathbf{K}_{img}^{ref}).
\end{aligned}
\end{equation}

\paragraph{Layer-Selective Application. }
A default application of our method across all attention layers can inadvertently disrupt the image generation process, leading to content artifacts and semantic inconsistencies. 
This is because layers in diffusion models are functionally specialized \cite{tumanyan2023plug, ding2024freecustom, wang2024instantstyle, frenkel2024implicit, qi2024deadiff}.
To ensure our method enhances style without corrupting content, we adopt a layer-selective strategy. We apply attention sharing only to a specific set of layers $\phi$ that are primarily responsible for generating style and fine-grained details. 
Through empirical validation (detailed in Section~\ref{sec:ablation}), we find that layers [19, 57) in Flux yield optimal performance.
Fig.~\ref{fig:block} demonstrates that this approach achieves strong style consistency across the collection of images while preserving both the naturalness of each image and its faithfulness to the corresponding text prompt.

\begin{figure*}[!t]
\vspace{0pt}
\centering
\includegraphics[width=1.00\textwidth]{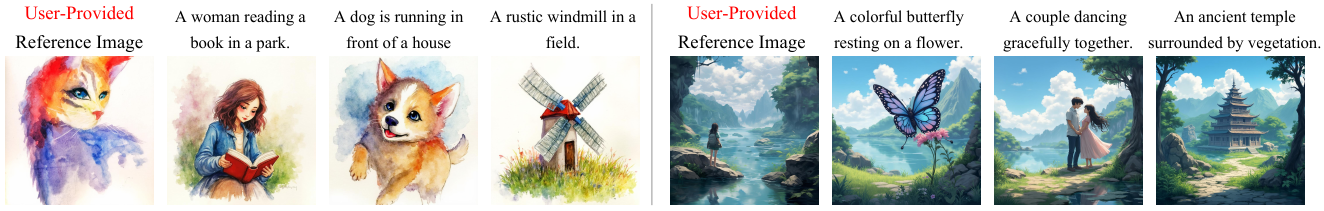}
\vspace{-12.5pt}
\caption{
\textbf{Generalization from Arbitrary Style Reference.}
Our method successfully generate diverse images conditioned on external, user-provided style references.}
\label{fig:seamless}
\vspace{0pt}
\end{figure*}

\subsection{Seamlessly Adapt to Any Style Reference}
A core principle of our method is the injection of style information via attention sharing, which is predicated on access to the $\mathbf{Q}$, $\mathbf{K}$, and $\mathbf{V}$ features from the style reference image. 
While these features are directly available if the reference is synthesized by the diffusion model itself, this assumption severely limits the method's applicability in real-world cases where users provide their own images. 
To bridge this critical gap, we propose a novel and simple feature extraction pipeline for arbitrary external images. 
This pipeline achieves this by directly leveraging the forward diffusion process. 
As shown in Fig.~\ref{fig:pipe} (d), for any given timestep t, we first add a noise to the reference image's latent representation. This procedure yields a noisy latent, which emulates the intermediate state during the standard generation process. 
By feeding this noisy latent into the diffusion model, we can query its intermediate layers to obtain the $\mathbf{Q}$, $\mathbf{K}$, and $\mathbf{V}$ features.
We showcase qualitative results in Fig.~\ref{fig:seamless} and provide pseudocode in the Appendix ~\ref{sec:feature_pipeline}.
% with further analysis

\section{Experiment}
\label{sec:exp}

\subsection{Implementation Detail}
\label{sec4:imple}
\paragraph{Settings.} We apply our method on Flux.1-dev~\citep{flux} by replacing the MM-Attention in [19,57) layers  with ShiftPE and AAS. 
For inference, we use Rectified Flow sampler with 30 sampling steps and set the classifier-free guidance scale to 3.5. Our evaluation dataset consists of 100 prompt sets from StyleAligned~\citep{hertz2024style}.
% These prompts cover a wide range of generation targets and a diverse set of style descriptions.

\paragraph{Metrics.} 
\textbf{(1) Text Alignment ($S_{text}$).} To evaluate how well the images match the prompts, we calculate the cosine similarity using CLIP~\citep{radford2021learning} between the images and prompts. 
\textbf{(2) Style Consistency ($S_{sty}$).} To evaluate the style consistency across generated images, we calculate the pairwise average cosine similarity of the CLIP embeddings within the set of generated images. 
\textbf{(3) DINO ($S_{dino}$).} Following ~\citep{voynov2023pplus, ruiz2023dreambooth, hertz2024style,li2025q}, we assess style consistency by computing the pairwise average cosine similarity of DINO~\citep{caron2021emerging, oquab2023dinov2} embeddings in the generated image set. This choice is motivated by the fact that CLIP, trained with category labels, often rates images with similar content but different styles as highly similar, DINO is trained in a self-supervised manner, better differentiates styles.

\begin{figure*}[!t]
\vspace{0pt}
\centering
\includegraphics[width=1.00\textwidth]{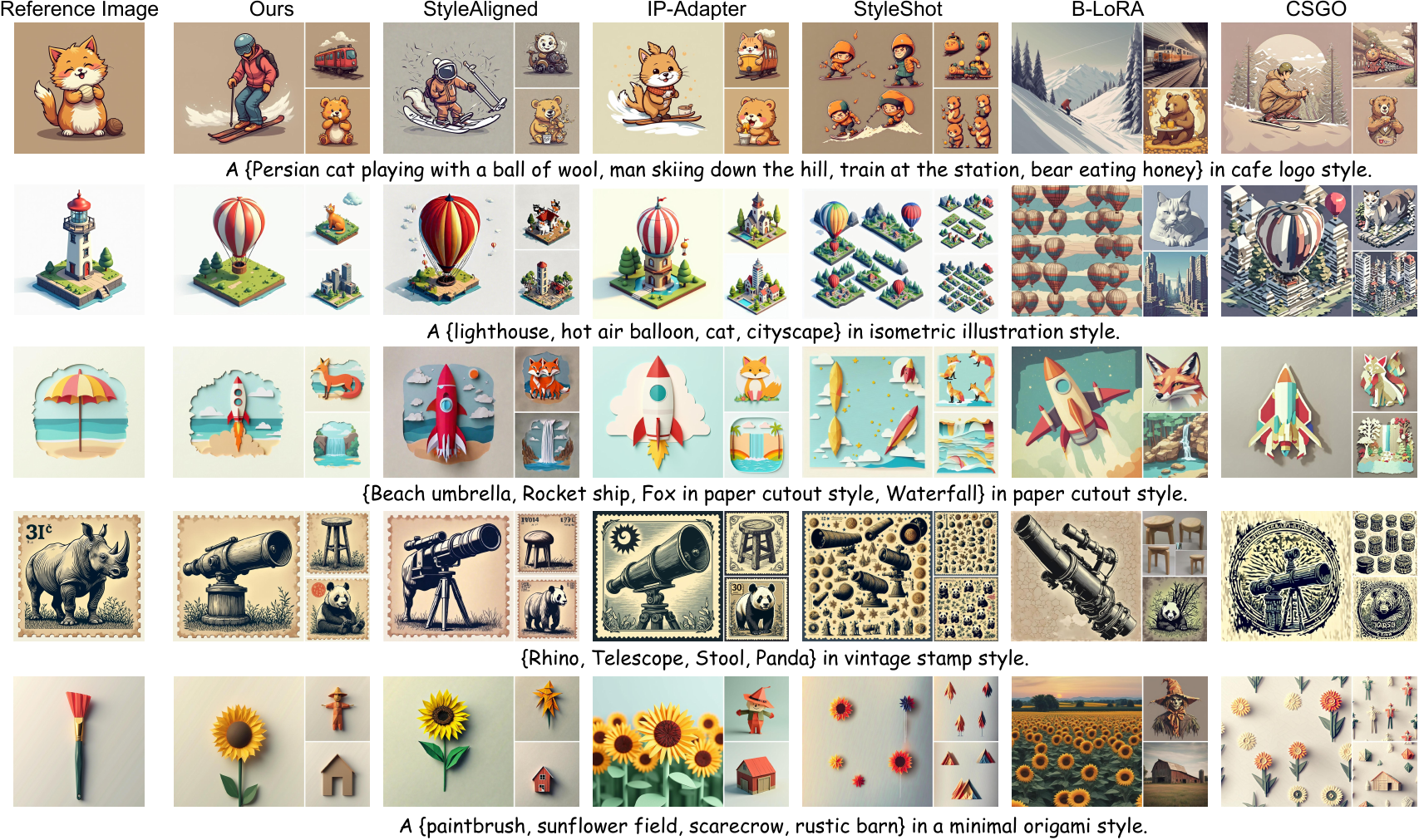}
\vspace{-15pt}
\caption{\textbf{Qualitative comparison between our method and other approaches.} Our method generates images with superior style consistency and accurate alignment with text prompts.}
\label{fig:compare}
\vspace{0pt}
\end{figure*}

\begin{table}[t]
\vspace{-5pt}
\caption{\textbf{Quantitative comparison.} We evaluate the generated image sets in terms of text alignment (\(S_{text}\)), style consistency (\(S_{sty}\) and \(S_{dino}\)). The symbols $\uparrow$ indicate higher values are better. Best result is marked in \textbf{bold}, and the second-best results are marked in \underline{underline}.}
\begin{center}
\footnotesize
\begin{tabular}
{c|c|cccccc}
\toprule
Metric & \multicolumn{1}{c}{Ours} & \multicolumn{1}{|c}{StyleAligned} & \multicolumn{1}{c}{IP-Adapter} & \multicolumn{1}{c}{B-Lora}  & \multicolumn{1}{c}{StyleShot} & \multicolumn{1}{c}{CSGO} \\
\midrule
   \(S_{text}\) $\uparrow$ & \textbf{0.282} & 0.277 & 0.278 & 0.259 & 0.276 & \underline{0.280} \\
  \(S_{sty}\) $\uparrow$ &  \textbf{0.740} & 0.728 & \underline{0.737} & 0.625 & 0.693 & 0.686 \\
  \( S_{dino}\) $\uparrow$ &  \textbf{0.554}& 0.512 & \underline{0.522}  & 0.271 & 0.511 & 0.450 \\
\bottomrule
\end{tabular}
\end{center}
\vspace{-17.5pt}
\label{table:comparison}
\end{table}

\subsection{Comparison}
\label{sec4:compare}

\paragraph{Competing Methods.}
We compare our method against five methods: StyleAligned~\citep{hertz2024style}, IP-Adapter~\citep{ye2023ip} (designed by the Instant-X~\citep{instantx} for Flux.1-dev), B-Lora~\citep{frenkel2024implicit}, StyleShot~\citep{gao2024styleshot}, CSGO~\citep{xing2024csgo}.

\paragraph{Qualitative Comparison.} 
Fig.~\ref{fig:compare} presents a qualitative comparison between our method and other approaches. 
IP-Adapter struggles to capture style effectively, failing to maintain consistent style in generated images and introducing content leakage issues. 
B-Lora, due to training its LoRA with only a single image, results in highly challenging training, consequently leading to generated images that struggle to capture the style of the reference image.
StyleAligned demonstrates improved style consistency, but the visual quality of its generated images still falls short compared to our method. StyleShot and CSGO frequently exhibit issues of image duplication and visual artifacts.
In contrast, our method produces images that not only exhibit superior style consistency across generated images but also perfectly align with the given prompts.
User Study can be found in Appendix~\ref{sec:user_study}.

\paragraph{Quantitative Comparison.} 
Table~\ref{table:comparison} presents the quantitative comparison results between our method and other style-aligned image generation approaches on the test set. Our method achieves the highest scores in terms of $S_{text}$, $S_{sty}$, and $S_{dino}$ metrics, demonstrating that it establishes the optimal balance between text-image alignment and style consistency among generated images.

\subsection{Ablation Study}
\label{sec:ablation}

% ShiftPE 无需 ablation 前面已经写的很清楚了
% \paragraph{Effect of ShiftPE.} 
% The experimental results in (k) and (m) of Table~\ref{table:ablation} demonstrate the impact of ShiftPE. When default RoPE used, the $S_{sty}$ and $S_{dino}$ metric rises to nearly 1, revealing a severe content leakage problem. Besides, Fig.~\ref{fig:ablation} illustrates that default RoPE cause the four generated images to become nearly identical, whereas ShiftedPE maintain style consistency while ensuring alignment with the respective prompts.

\begin{table}[t]
\vspace{0pt}
\caption{\textbf{Ablation study.} The Layer and $\lambda$ denote the ablation study for Layer-Selective Application and scaling factor $\lambda$, respectively. The ``w/o SAS'' represent the ablation study for the SAS module. }
\begin{center}
\footnotesize
\setlength{\tabcolsep}{2.60mm}{
\begin{tabular}
{c|c|c|ccc|ccc}
\toprule
\multicolumn{1}{c}{Type} & \multicolumn{1}{|c}{\#} & \multicolumn{1}{|c}{$\lambda$} & \multicolumn{1}{|c}{0-19} & \multicolumn{1}{c}{19-38} & \multicolumn{1}{c}{38-57} & \multicolumn{1}{|c}{\(S_{text}\) $\uparrow$} & \multicolumn{1}{c}{\(S_{sty}\) $\uparrow$} & \multicolumn{1}{c}{\(S_{dino}\) $\uparrow$} \\
\midrule
{Original} & - & - & - & - & - & {0.284} & {0.658} & {0.313} \\
\midrule 
\multirow{5}{*}{Layer} & (a) & 1.10 & $\checkmark$ & $\times$ & $\times$ & {0.283} & {0.669} & {0.323} \\
 & (b)  & 1.10 & $\times$ & $\checkmark$ & $\times$ & {0.283} & {0.697} & {0.439} \\
 & (c)  & 1.10 & $\times$ & $\times$ & $\checkmark$ & {0.286} & {0.682} & {0.368} \\
 & (d)  & 1.10 & $\checkmark$ & $\checkmark$ & $\times$ & {0.274} & {0.697} & {0.420} \\
 & (e) & 1.10 & $\checkmark$ & $\checkmark$ & $\checkmark$ & {0.273} & {0.735} & {0.531} \\
\midrule
\multirow{5}{*}{$\lambda$} & (f) & 0.90 & $\times$ & $\checkmark$ & $\checkmark$ & {0.286} & {0.694} & {0.435} \\
& (g) & 0.95 & $\times$ & $\checkmark$ & $\checkmark$ & {0.285} & {0.703} & {0.460} \\
& (h) & 1.00 & $\times$ & $\checkmark$ & $\checkmark$ & {0.286} & {0.711} & {0.485} \\
& (i) & 1.05 & $\times$ & $\checkmark$ & $\checkmark$ & {0.285} & {0.725} & {0.522} \\
& (j) & 1.15 & $\times$ & $\checkmark$ & $\checkmark$ & {0.274} & {0.755} & {0.561} \\
\midrule
w/o SAS & (k) & 1.10 & $\times$ & $\checkmark$ & $\checkmark$ & {0.281} & {0.733} & {0.525} \\
\midrule
\textbf{Ours} & (l) & 1.10 & $\times$ & $\checkmark$ & $\checkmark$ & {0.282} & {0.740} & {0.554} \\
\bottomrule
\end{tabular}
}
\end{center}
\vspace{-12.5pt}
\label{table:ablation}
\end{table}

\begin{figure*}[!t]
\vspace{0pt}
\centering
\includegraphics[width=1.00\textwidth]{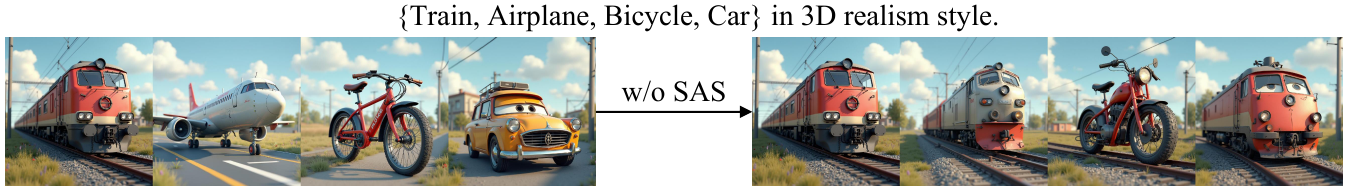}
\vspace{-12.5pt}
\caption{Visualization of Ablation Study on SSA.}
\label{fig:ablation}
\vspace{-5pt}
\end{figure*}

\paragraph{Effect of Selective Attention Sharing (SAS).} 
Table~\ref{table:ablation} (k) and (l) show results without and with the SAS.
Omitting the SAS reduces $S_{sty}$ and $S_{dino}$ but does not improve $S_{text}$. 
Fig.~\ref{fig:ablation} displays outputs under both configurations. Without the SAS, the second image mismatch its textual descriptions (airplane), and style similarity decreases. 

\paragraph{Effect of Layer-Selective Application.} 
% We divided the blocks in Flux.1-dev into three groups: Group 1 contains 19 Double Blocks, Group 2 has the first 19 Single Blocks, and Group 3 includes the remaining 19 Single Blocks. The Ablation-Block part in Table~\ref{table:ablation} shows quantitative results of applying SSA layers in different blocks.
% Comparing settings (a), (b), and (c) shows that Group 2 blocks contain the richest style information.
% Comparing settings (b) and (d) reveals that using SSA layer in Group 1 blocks does not improve style consistency and even has a negative effect, which is also confirmed by comparing settings (e) and (m). 
% Comparing settings (d), (e), and (m) demonstrates that applying SSA layers in the last two groups achieves a good balance between image-to-prompt alignment and style consistency across images.
% Finally, we apply our method exclusively in
% the Single Blocks (Group 2 and Group 3) to enhance style consistency among generated images.
To find the optimal placement for our method, we divide the Flux.1-dev blocks into three distinct groups: Group 1 (19 Double Blocks), Group 2 (the first 19 Single Blocks), and Group 3 (the remaining 19 Single Blocks). 
As detailed in Table~\ref{table:ablation}, ablation study reveals several insights. 
First, comparing settings (a), (b), (c) highlights that Group 2 is most critical for capturing style information. 
Conversely, applying our method to Group 1 is ineffective and even detrimental to style consistency, a finding supported by two comparisons: (b), (d) and (e), (l).
Finally, settings (d), (e), and (l) demonstrate that integrating our method across both Groups 2 and 3 strikes the best balance between text-image alignment and style consistency. 
Based on these findings, our final design applies our method exclusively to all Single Blocks (Groups 2 and 3).

\paragraph{Effect of scaling factor.} 
Table~\ref{table:ablation} presents the quantitative results of our method for various values of $\lambda$. 
Specifically, as $\lambda$ increases, the style consistency ($S_{sty}$, $S_{dino}$) improves, but this comes at the drop of $S_{text}$. 
Our experiments indicate that, generally, setting the strength parameter $\lambda$ to 1.1 strikes a favorable balance between style consistency and text alignment. 
% Figure~\ref{fig:lambda} illustrates how the target image progressively aligns with the style of the reference image as $\lambda$ increases. 

\subsection{Generalization and Compatibility}
\label{sec4:additional}

\paragraph{Architectural Generalization.}
To demonstrate our method is not tailored to Flux, we test its performance on other DiT models, including SD3 and SD3.5.
As shown in Fig.~\ref{fig:sd3}, our method seamlessly integrates with these backbones.
More results can be found in Appendix ~\ref{sec:more}.
% proving its broad applicability and robustness.

\paragraph{Compatibility with Existing Tools.} 
A key advantage of our method is its training-free, plug-and-play design. 
This allows for effortless composition with other generation techniques. 
We showcase this in Appendix ~\ref{sec:more} by combining our method with techniques like FluxControl and Dreambooth. 
% We showcase this in Fig.~\ref{fig:dreambooth} by combining our method with other methods like FluxControl and Dreambooth. 
% These results highlight its utility as a versatile module for generative workflows.

\begin{figure*}[!t]
\centering
\includegraphics[width=1.00\textwidth]{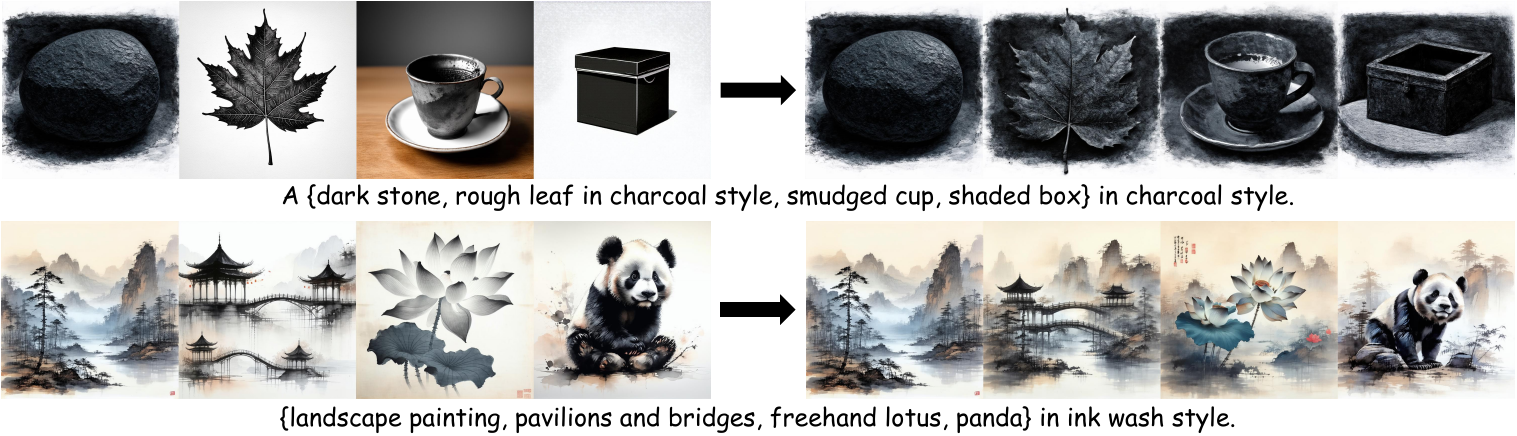}
\vspace{-12.5pt}
\caption{Generation results of applying AlignedGen to other models. (Top) SD3, (Bottom) SD3.5.}
\label{fig:sd3}
\vspace{-5pt}
\end{figure*}

\section{Conclusion}
\label{sec:conclusion}
% In this paper, we propose StoryDiffusion, a novel method that can generate consistent images in a training-free manner for storytelling and transition these consistent images into videos. Our Consistent Self-Attention builds connections among multiple images to efficiently generate images with consistent faces and clothing. We further propose the Semantic Motion Predictor to transition these images into videos and better narrate the story. We hope that our StoryDiffusion can inspire future controllable image and video generation endeavors.

% 本文提出了 AlignedGen，一种无需训练的 style-aligned image generation method，可以使用在基于MM-DiT扩散模型中。我们提出的 Shifted Position Embedding 通过引入不重叠的位置索引极大减轻了先前方法应用于 MM-DiT 扩散模型中所带来的内容泄露问题；SSA layer和Delicate Attention Replacement则进一步提升了生成图像间的风格一致性。大量实验证明了我们方法的有效性和优越性——可以在有效增强生成图像间风格一致性的同时保持良好的文本可控性。我们希望AlignedGen可以为未来的style-aligned image generation相关工作带来启发。
In this paper, we propose AlignedGen, a training-free style-aligned image generation method for Flux.
Our Shifted Position Embedding significantly mitigates the text controllability degradation and content leakage issue observed in prior methods when applied to Flux through the introduction of non-overlapping positional indices for reference image and target images. 
The Selective Shared Attention layer and Delicate Attention Replacement further enhance style consistency among generated images. 
Extensive experiments demonstrate the effectiveness and superiority of our approach - it substantially improves inter-image style consistency while maintaining robust text controllability.
Our method exhibits plug-and-play characteristics, enabling seamless integration with other controllable generation approaches.
Moreover, our method can also be applied to other MM-DiT diffusion models.
We hope AlignedGen can inspire future research in style-aligned image generation methods.

\clearpage
\bibliographystyle{plain}
\bibliography{ref}

%%%%%%%%%%%%%%%%%%%%%%%%%%%%%%%%%%%%%%%%%%%%%%%%%%%%%%%%%%%%
\clearpage
\appendix

\section{Feature Extraction Pipeline}
\label{sec:feature_pipeline}

\begin{algorithm}[H]
\caption{Algorithm for Caching Q, K, V from a User-Provided Reference Image}
\label{alg:ref_image_cache}
\begin{algorithmic}[1]
\Require User-provided reference image $I$, number of inference timesteps $T$
\Ensure Cached attention Q, K, V pairs: $\text{cached\_qkv}$

\State Initialize a random noise: $\text{noise} \sim \mathcal{N}(0, 1)$
\State $\text{latent} \gets \text{vae\_encode}(I)$
\State $\text{cached\_qkv} \gets \{\}$ \Comment{Initialize an empty dictionary}
\For{$t \gets T$ \textbf{down to} $0$} \Comment{Iterate over $T$ timesteps in reverse}
    \State $\text{noise\_input} \gets \frac{t}{T} \cdot \text{noise} + (1 - \frac{t}{T}) \cdot \text{latent}$ \Comment{A standard interpolation forward}
    \State $Q_t, K_t, V_t \gets \text{DiT}(\text{noise\_input}, t, \text{others})$ \Comment{e.g., prompt embeds}
    \State $\text{cached\_qkv}[t] \gets (Q_t, K_t, V_t)$
\EndFor
\State \Return $\text{cached\_qkv}$
\end{algorithmic}
\end{algorithm}

\section{User Study}
\label{sec:user_study}

We conducted a user study to assess the results of our method compared to other methods. 
In each question, participants were asked to rank the methods according to the style consistency among their generated images and the alignment of these images with the prompts.
In total, our user study involved 40 participants, and Table~\ref{table:user} shows the result.
Our method significantly outperforms competing approaches, achieving an average ranking of 1.20, which demonstrates its superior performance.

\begin{table*}[h]
\caption{\textbf{User Study.} The results represent the average ranking outcomes for each method (lower is better). Our approach significantly outperformed other comparative methods in the user study.}
\begin{center}
\footnotesize
\begin{tabular}
{cccccc}
\toprule
Ours & StyleAligned & IP-Adapter & B-Lora & StyleShot & CSGO \\
\midrule
\textbf{1.20} & 3.66 & 2.04 & 4.76 & 4.09 & 5.25 \\
\bottomrule
\end{tabular}
% }
\label{table:user}
\end{center}
\end{table*}
\vspace{0pt}

\section{Additional Compare With Image Editing Methods}
\begin{figure*}[h]
\vspace{0pt}
\centering
\includegraphics[width=1.00\textwidth]{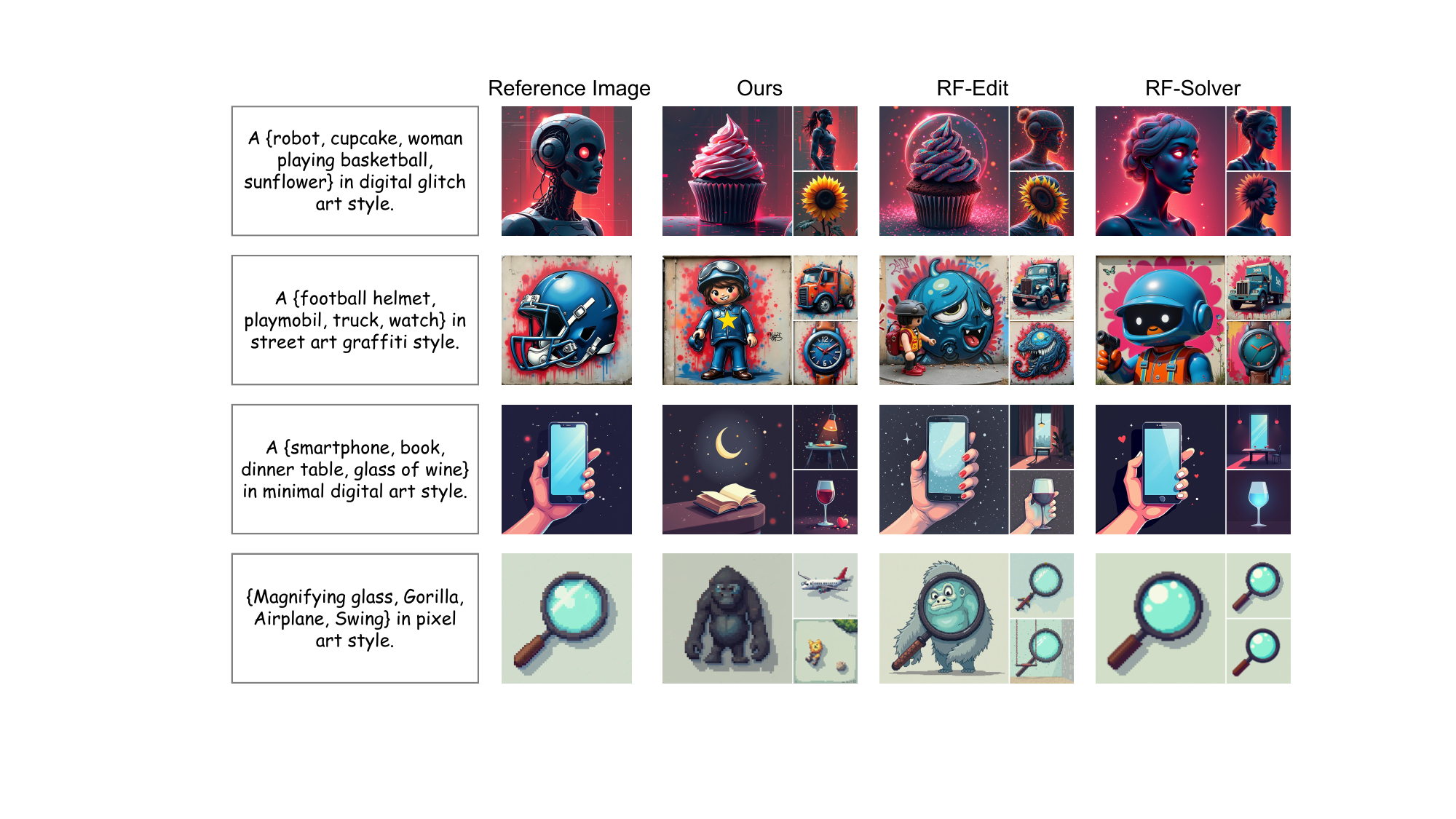}
\vspace{0pt}
\caption{Qualitative comparison between our method and other zero-shot image editing methods.}
\label{fig:supp-compare}
\vspace{0pt}
\end{figure*}

Some zero-shot image editing methods can also be applied to style-aligned image generation tasks. We selected two Flux-based editing methods for comparison: RF-Solver~\citep{wang2024taming} and RF-Edit~\citep{rout2024semantic}.
Fig.~\ref{fig:supp-compare} presents qualitative comparison results between our method and editing methods. Image editing methods often exhibit insufficient editing strength when handling significant image edits such as modifying the main content of the image, leading to persistent retention of original content in the edited results.
Table~\ref{table:edit} presents the quantitative comparison results between our method and these two methods. As can be observed, our method achieves significantly higher scores on $S_{text}$, $S_{sty}$, and $S_{dino}$ compared to these two baseline methods.
It is worth noting that zero-shot editing methods, limited by the instability of editing processes, often require meticulous parameter adjustments to achieve satisfactory results, which limits their practical applicability. In contrast, our method exhibits strong robustness and can typically achieve favorable outcomes in most scenarios without necessitating fine-tuning.

\begin{table*}[h]
\caption{\textbf{Quantitative comparison with zero-shot image editing methods.} Best result is marked in \textbf{bold}.}
\begin{center}
\footnotesize
% \resizebox{\textwidth}{!}{
\setlength{\tabcolsep}{3.5mm}{
\begin{tabular}
{c|c|cc}
\toprule
Metric & \multicolumn{1}{c}{Ours} & \multicolumn{1}{|c}{RF-Solver} & \multicolumn{1}{c}{RF-Edit} \\
\midrule
   \(S_{text}\) $\uparrow$ & \textbf{0.282} & 0.278 & 0.280 \\
  \(S_{sty}\) $\uparrow$ &  \textbf{0.740} & 0.703 & 0.701 \\
  \( S_{dino}\) $\uparrow$ &  \textbf{0.554} & 0.487 & 0.456 \\
\bottomrule
\end{tabular}
}
\label{table:edit}
\end{center}
\end{table*}
\vspace{0pt}

\section{More Visual Results}
\label{sec:more}

\begin{figure*}[!t]
\vspace{0pt}
\centering
\includegraphics[width=1.00\textwidth]{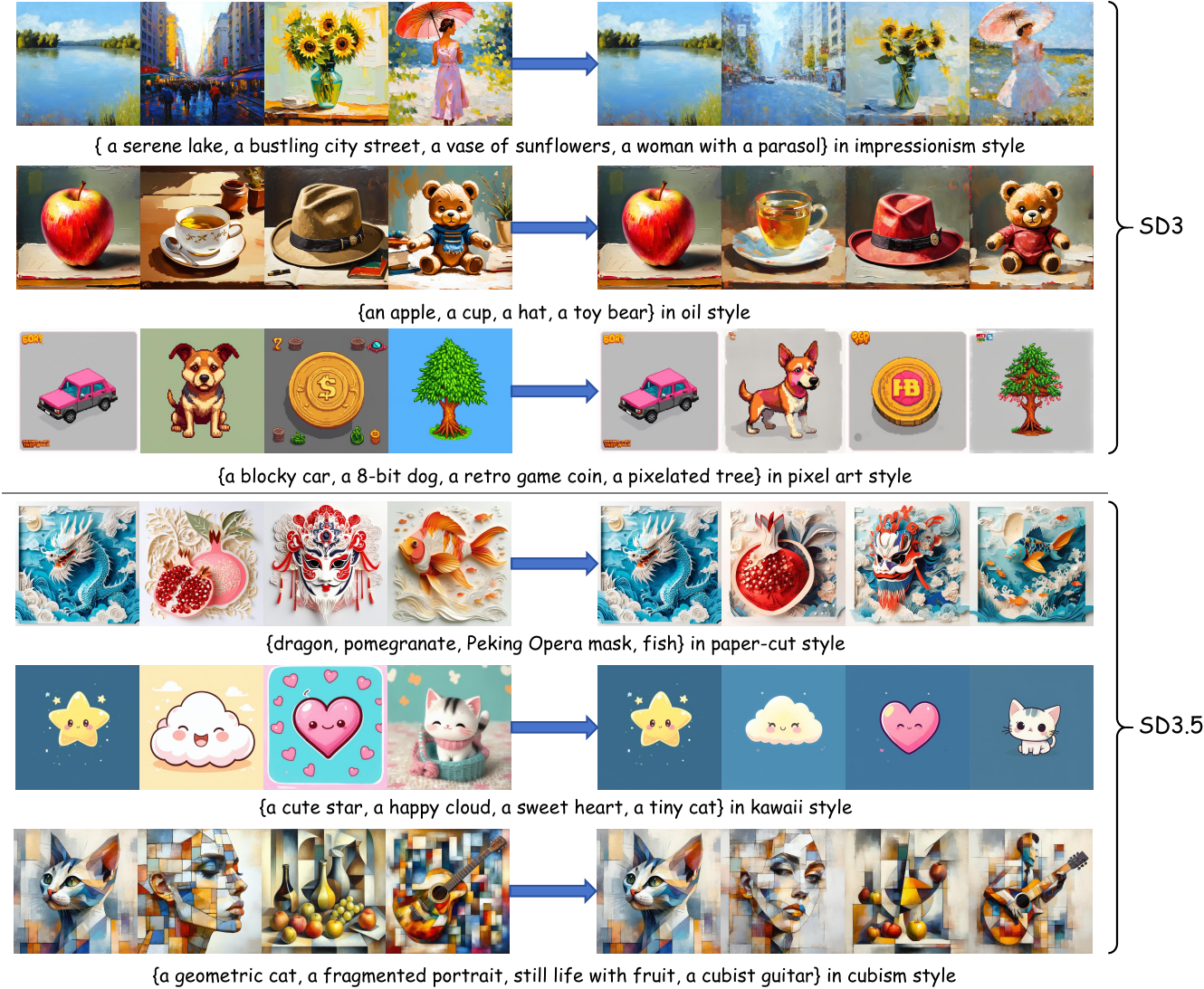}
\vspace{0pt}
\caption{Generation results of applying AlignedGen to other DiT models. The left side presents images generated by the original model, while the right side displays images generated after applying our method.}
\label{fig:supp-sd}
\vspace{0pt}
\end{figure*}

\begin{figure*}[!t]
\vspace{0pt}
\centering
\includegraphics[width=1.00\textwidth]{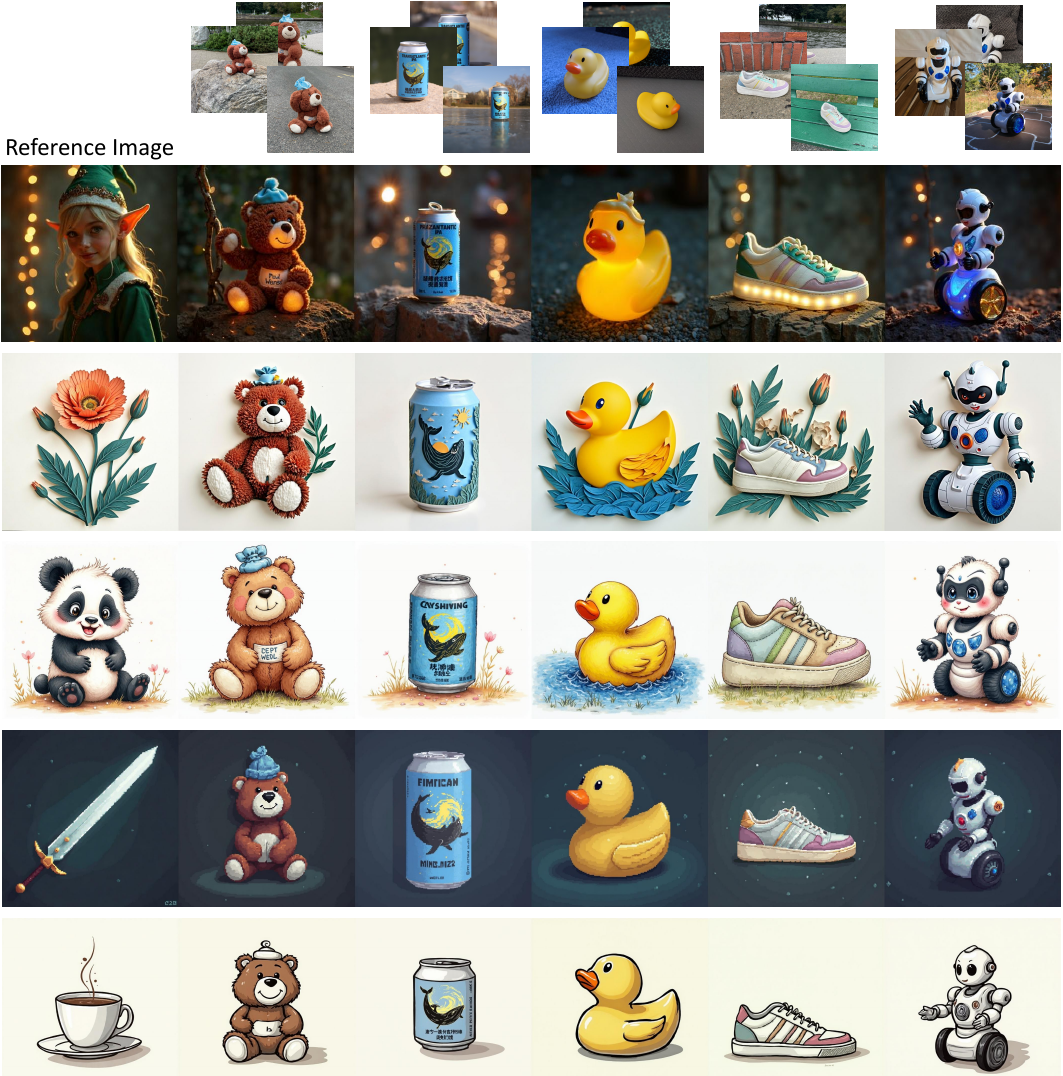}
\vspace{0pt}
\caption{Subject-driven image generation with AlignedGen. Each row shows style aligned image set using the reference image on the left,
applied on different personalized diffusion models, fine-tuned over the personalized content on top.}
\label{fig:supp-dream}
\vspace{0pt}
\end{figure*}

\begin{figure*}[!t]
\vspace{0pt}
\centering
\includegraphics[width=0.80\textwidth]{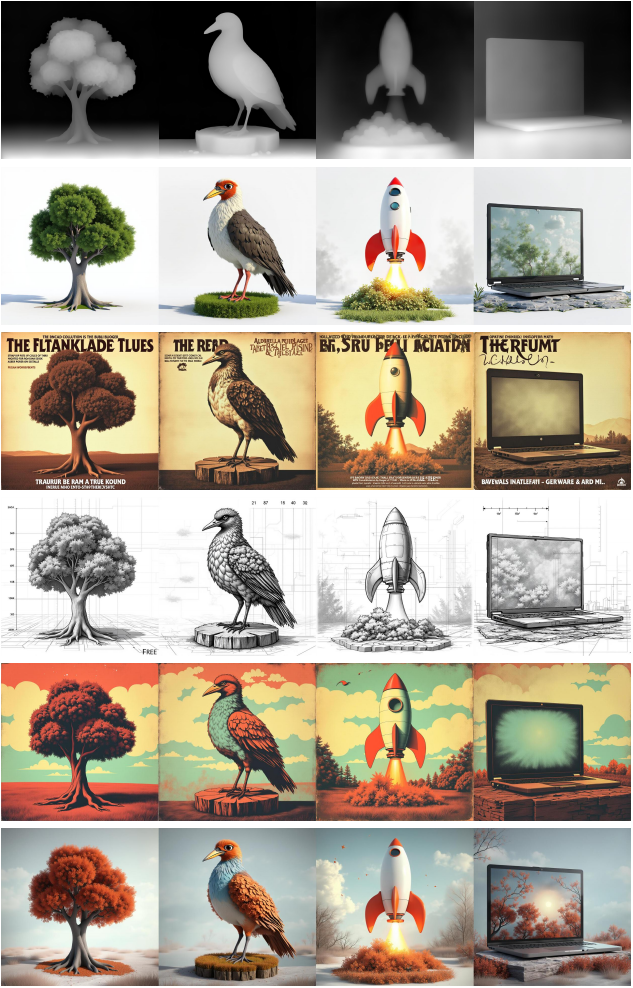}
\vspace{0pt}
\caption{Depth Control with AlignedGen.}
\label{fig:supp-control}
\vspace{0pt}
\end{figure*}

% 图~\ref{fig:supp-sd}展示了将AlignedGen应用在其余MM-DiT架构扩散模型上的生成结果。图~\ref{fig:supp-dream}展示了AlignedGen和Dreambooth结合的生成结果。图~\ref{fig:supp-control}展示了将depth control和AlignedGen结合的生成结果。
Fig.~\ref{fig:supp-sd} presents the generation results of applying AlignedGen to other MM-DiT architecture diffusion models. Fig.~\ref{fig:supp-dream} demonstrates the generation outcomes from combining AlignedGen with Dreambooth. Fig.~\ref{fig:supp-control} illustrates the generation results achieved by integrating depth control with AlignedGen.

\end{document}